\documentclass[review]{elsarticle}
\usepackage{amssymb}
\usepackage{graphicx}
\usepackage{amsthm}
\usepackage{amsmath}
\usepackage[linesnumbered,ruled]{algorithm2e}
\usepackage{natbib}
\usepackage{booktabs}
\usepackage{mathrsfs}
\usepackage{subfigure}

\journal{ArXiv}

\begin{document}

\begin{frontmatter}

\title{Hyper-parameter optimization based on soft actor critic and hierarchical mixture regularization}

\author{Chaoyue Liu, Yulai Zhang}
%\author{Yulai Zhang$^{a}$, Yaru Li$^{a}$, Guiming Luo$^b$ }
\address{School of Information Technology and Electronics Engineering, \\ Zhejiang University of Science and Technology, Hangzhou, China, 310023, \\ zhangyulai@zust.edu.cn}
%\address[b]{School of Software, Tsinghua University, Beijing, China, 100084, gluo@tsinghua.edu.cn}
\begin{abstract}
Hyper-parameter optimization is a crucial problem in machine learning as it aims to achieve the state-of-the-art performance in any model. Great efforts have been made in this field, such as random search, grid search, Bayesian optimization. In this paper, we model hyper-parameter optimization  process as a Markov decision process, and tackle it with reinforcement learning. A novel hyper-parameter optimization method based on soft actor critic and hierarchical mixture regularization has been proposed. %In order to speed up the convergence, we proposed hierarchical mixture regularization. 
Experiments show that the proposed method can obtain better hyper-parameters in a shorter time.

\end{abstract}


\end{frontmatter}

\section{INTRODUCTION}
Hyper-parameter optimization is a key problem in machine learning. Its goal is to achieve the best performance in any model by selecting the appropriate hyper-parameters. The hyper-parameters are different from the internal parameters. They are specified before model training and will not be changed in the training process \cite{wu2020efficient}. Hyper-parameter optimization is a ticklish and time-consuming problem, especially on complex models or on large datasets. Moreover, manually setting hyper-parameters is a professional task, which requires researchers to have a deep understanding of the models. To tackle this problem, we have to confront several challenges. First, the relationship between hyper-parameters and their performance cannot be clearly expressed. So we cannot utilize conventional method like gradient-based method to solve it. Second, the search space of each hyper-parameter is so large that the whole search space expands exponentially.

In this paper, we proposed a new method combining soft actor critic with hierarchical mixture regularization to solve the hyper-parameter optimization problem \cite{wang2020improving}. Specifically, the hyper-parameter optimization process is regarded as a Markov decision process and reinforcement learning method is used to tackle it. A LSTM (Long Short-term Memory Network) is used as the environment, which means the action of the reinforcement model is input to LSTM to obtain the value of next state. The actions are obtained from the actor network constructed by a MLP (Multi-layer Perceptron). In addition, the expectation of the reward is calculated by a network named critic. The final objective is to maximize the expected reward which is related to the performance of the hyper-parameters. In other words, we aim to maximize the expected discounted reward from the loss of the validation dataset. As we know, in contrast to conventional method, reinforcement learning method has relatively slow convergence rate. To solve this problem, we make use of the hierarchical mixture regularization to get optimal hyper-parameters faster.

\section{Related Work}
An algorithm that optimizes the hyper-parameters by DQN (Deep Q-Learning Network) has been proposed in \cite{hansen2016using}. This paper describes how to define the environment, state, action and reward function in the problem of hyper-parameter optimization. %To the best of our knowledge, it is the first paper that combined hyper-parameter optimization problem and reinforcement learning.

Neural architecture search problem is similar to our study, which aims to search the best structure of neural network automatically. Google Brain uses deep reinforcement learning to search the structures of the deep neural network on a given dataset \cite{zoph2016neural}, where policy gradient method is adopted to maximize the expected accuracy of the sampled architectures \cite{sutton2000policy}. EAS (Efficient Architecture Search) has been proposed in \cite{cai2018efficient}, which is a new framework that is able to utilize knowledge stored in previously trained networks and take advantage of the existing successful architectures in the target task to explore the architecture space efficiently. EAS utilizes a reinforcement learning agent as the meta-controller. 

An algorithm for optimizing the learning rate of the SGD (Stochastic Gradient Descent) within the actor-critic framework on image classification datasets has been proposed in \cite{xu2017reinforcement}. The experiments have demonstrated their method can successfully adjust learning rate for different datasets with CNN model structures and can reduce oscillations during training. The work in \cite{jomaa2019hyp} provides empirical evidence of the  feasibility of optimizing hyper-parameters based on reinforcement learning and a novel policy based on Q-learning method is proposed to explore high-dimensional hyper-parameter spaces in this paper. In \cite{jia2019rpr} a method using real experience and predictive reward alternately has been proposed, it aims to accelerate the training process by predicting rewards.% generated from a prediction network . 
 
\section{PRELIMINARIES}
\subsection{Hyper-parameter Optimization}\label{cl}
The purpose of hyper-parameter optimization is to find a set of hyper-parameters that obtain the best performance in the whole hyper-parameter space at a reasonable cost. Hyper-parameter optimization can be formally defined as follows.

We denote $\mathcal{A}$ as a machine learning algorithm, $\mathit{\Lambda}$ as the hyper-parameter space, $\mathcal{D}$ as the set of all datasets, $\mathcal{A} _{\lambda } $ as the machine learning algorithm with its hyper-parameters $\lambda$ where $\lambda \in \mathit{\Lambda}$. The hyper-parameter space $\mathit{\Lambda}$ can include continuous or discrete hyper-parameters, and its dimension can be arbitrary. The goal of hyper-parameter optimization is to identify the optimal hyper-parameter set $\lambda ^{*} $ that minimizes the loss function denoted as $\mathcal{L}$, so the equation can be written as follows:
\begin{equation}
\lambda ^{*}=argmin _{\lambda\in \mathit{\Lambda } } \mathbb{E}_{(D_{tr},D_{va} )\sim \mathcal{D} }\mathcal{L}(\mathcal{A}_{\lambda }(D_{tr}),D_{va})
\end{equation}
where both $D_{tr}$ as the training dataset and $D_{va}$ as the validation set are sampled from $\mathcal{D}$. Considering the machine learning algorithm as a black box, we cannot clearly analyze the importance of each hyper-parameter. And with the increase of the number of hyper-parameters, it is difficult to find the globally optimal hyper-parameters \cite{jomaa2019hyp}.

So far, there are diverse methods to solve hyper-parameter optimization problems. Such as grid search or random search, it is widely used to improve performance, but these methods are only suitable for small hyper-parameter space. With the enlargement of hyper-parameter space, the performance of optimization becomes unstable and fluctuates drastically. In addition, grid search is affected by the number of hyper-parameters and the random search is inefficient due to lack of guidance.

Bayesian optimization method has achieved success at limited cost, which is a sequential model-based optimization method \cite{hutter2011sequential}. Bayesian optimization requires probabilistic surrogate model and acquisition function. The acquisition function maintains a balance between exploration and exploitation, which is a bit like the concept of reinforcement learning. Selecting hyper-parameters from an already well-explored area or a previously unexplored area is a difficult trade-off. The former may take less time but reach the local optimum, while the latter is more likely to find the global optimum. There is also a kind of hyper-parameter optimization algorithm based on gradient, which implement hyper-parameter optimization via calculating the partial derivative of the loss function \cite{maclaurin2015gradient,pedregosa2016hyperparameter}.

On the whole, we can model hyper-parameter optimization as reinforcement learning problem since hyper-parameter optimization is a sequential decision process and has a explicit objective that minimizes the loss function.
\subsection{Reinforcement Learning}
The Markov decision process, which is the basic concept in reinforcement learning,is defined by a tuple $\mathcal{M} = (\mathcal{S} ,\mathcal{A} ,\mathcal{R} ,\mathcal{P} ,\gamma )$, with the following ingredients: states $s \in \mathcal{S}$ called state space, actions $a \in \mathcal{A}$ as the set of all actions, $\mathcal{R}$ as the reward function $\mathcal{S} \times \mathcal{A} \to \mathbb{R}$, $\mathcal{P}$ as the transition function $\mathcal{S} \times \mathcal{A} \to \mathcal{S}$, $\gamma$ as the discount factor in $[0,1]$ that is used to calculate expected reward $R={\textstyle \sum_{t}^{T}} \gamma^{t} r_{t} $\cite{sutton2018reinforcement}. An agent managed by a specified policy denoted as $\pi $ interacts with the environment and selects an action under a certain state. Then the environment returns the immediate reward and the next state to compute action value. The action value can be expressed as follows:
\begin{equation}
\mathrm {Q}   _{\pi } (s,a)=\mathbb {E}_{\pi } \left [  \sum_{t=0}^{\infty} \gamma^{t} r_{t} \mid S_{0}=s,A_{0}=a\right ]
\end{equation}
To obtain the optimal policy that maximizes the discounted cumulative reward, $\pi _{*} (s)\in argmax_{a} Q_{*}(s,a)$, there is a well-known Bellman equation as follows:
\begin{equation}
Q_{*}(s,a)=\mathbb{E} _{\pi}[r+\gamma max_{a^{\prime } }Q_{*}(s^{\prime } ,a^{\prime } )\mid S_{0}=s,A_{0}=a ]
\end{equation}

Reinforcement Learning has achieved great success in the past decade and has shown great potential in many fields, especially in the fields of robots and games, such as Go \cite{silver2016mastering,silver2017mastering2}, Chess, Shogi \cite{silver2017mastering} and Atari \cite{mnih2015human} etc. However, reinforcement learning has the problem of large sample demand, which requires thousands of samples for training, so it is not suitable for fields that are difficult to sample. According to whether the agent interacting with the environment is the same as the trained agent, reinforcement learning algorithm can be divided into on-policy and off-policy algorithm. On-policy algorithms,such as TRPO (Trust Region Policy Optimization), DDPG (Deep Deterministic Policy Gradient) or PPO (Proximal Policy Optimization), have the same behavior policy and target policy,  so the past experience cannot be reused \cite{schulman2015trust,lillicrap2015continuous,schulman2017proximal}. Off-policy algorithms aim to reuse past experience to reduce dependence on the number of samples, but they perform slightly worse in convergence and stability.

\section{ALGORITHM}
In this section, we will introduce our method in detail, including how to design reward signals, how to interact with the environment, and how to model hyper-parameter optimization problem as Markov decision process.

\subsection{MDP Formulation}
We can formulate hyper-parameter optimization process as a Markov decision process and define $f$ as the hyper-parameter response surface. The mapping can be written as follows:
\begin{equation}
f:\mathcal{D} \times \mathit{\Lambda } \to \mathbb{R}
\end{equation}
where $\mathit{\Lambda }$ is the hyper-parameter space mentioned previously. The action space is the same as the hyper-parameter space. Without loss of generality, we use validation loss of the model $\mathcal{A} _{\lambda } $ as the response. The equation can be expressed as follows:
\begin{equation}
f(D,\lambda)=\mathcal{L}(\mathcal{A}_{\lambda}(D^{tr} ),D^{va})
\end{equation}
Then we need datasets and hyper-parameters to calculate rewards. The reward function is described as follows:
\begin{equation}
R(D,\lambda)=\frac{1}{f(D,\lambda)-baseline}
\end{equation}
where $f$ is the validation loss function and the baseline is less than the minimum loss so that the reward function is strictly positive and the reward is magnified as much as possible. And we use LSTM as transition function to produce new state. On account of the characteristic of recurrent neural network, the new state has relevance with all previous states. The number of interactions is set to a fixed number. We set it as a small number so as to encourage agent to achieve optimal value in a controllable time.

\subsection{Mixture Regularization}
Mixture regularization method is a data augmentation technology to improve generalization ability of agents \cite{laskin2020reinforcement}. It combines two observations randomly sampled from the replay buffer and trains agent with the corresponding hybrid rewards. Data augmentation technology has been proved to effectively improve performance of algorithms \cite{cobbe2019quantifying,lee2019network}. This method can effectively increase the diversity of training data. The agent is more likely to have better generalization ability and faster learning speed. Specifically,mixture regularization method generates each augmented observation $\tilde{S}$ and $\tilde{r}$ by combining two observations $S_{i},S_{j}$ and their associated rewards $r_{i},r_{j}$. The equations can be written as follows:
\begin{equation}
\tilde{S}=\alpha S_{i}+(1-\alpha)S_{j}
\end{equation}
\begin{equation}
\tilde{r}=\alpha r_{i}+(1-\alpha)r_{j}
\end{equation}
where $\alpha$ is a hyper-parameter set to a positive number close to zero. The $S_{i}$ and $r_{i}$ are the current state and reward, whereas the $S_{j}$ and $r_{j}$ are randomly sampled from replay buffer. Then we proposed hierarchical mixture regularization to achieve better sample efficiency. So the equations change as follows:
\begin{equation}
\tilde{S}=(1-\alpha)^{n} S_{i}+ {\textstyle \sum_{j=1}^{n}} \alpha(1-\alpha)^{n-j}S_{j}
\end{equation}
\begin{equation}
\tilde{r}=(1-\alpha)^{n} r_{i}+ {\textstyle \sum_{j=1}^{n}} \alpha(1-\alpha)^{n-j}r_{j}
\end{equation}
where parameter $n$ decides the number of combined results. It is set to a small number, which ensures the combined results are not too far from $S_{i}$ and $r_{i}$.

\subsection{Model Architecture}
First, in order to make the network partially remember the previous states, we use LSTM to build the environment. To improve stability, the initial values of network are all one instead of sampling from the probability distribution. The size of hidden state and cell state should be small, which can help agent learn faster. The network is only used to generate the state of the next step. The interaction process is shown in Figure 1. The MLP is used as policy network.

\begin{figure}[!htbp]
	\includegraphics[width=380pt, bb=140 300 500 500,clip]{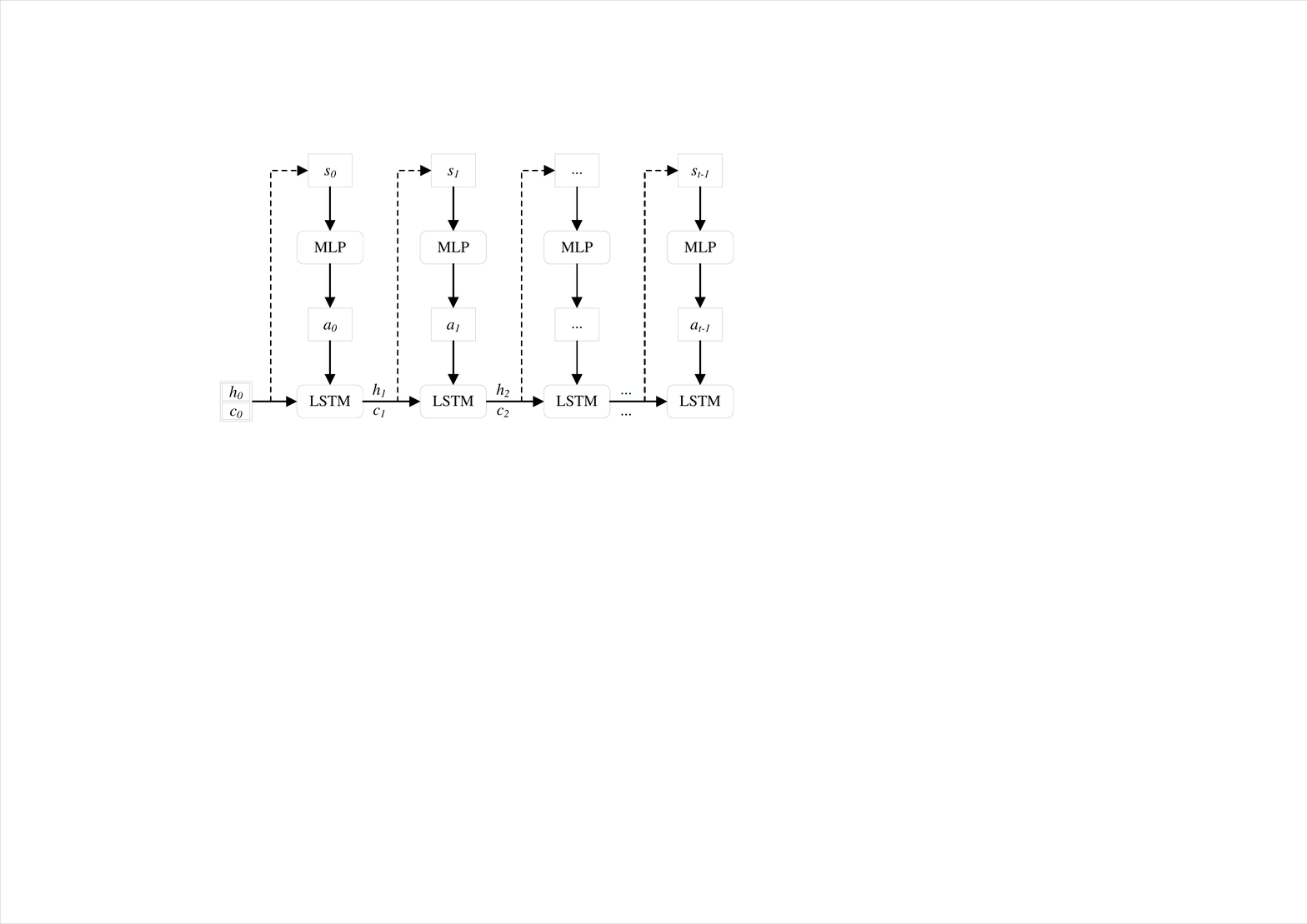} \centering \\
	\caption{Interaction process}
\end{figure}

Secondly, we build the network architecture on the basis of soft actor critic method. There are two state value networks, two action value networks and a policy network \cite{haarnoja2018soft}. The advantage of setting two action value networks separately is to improve the training stability and avoid overestimation by selecting a smaller action value \cite{van2016deep}. The objective is improved by adding the expected entropy of policy $\pi$. The improved objective can be written as follows:
\begin{equation}
J(\pi)= {\textstyle \sum_{t=0}^{T}}\mathbb{E}_{(s_{t},a_{t})\sim \pi} \left [ r(s_{t},a_{t})+\alpha \mathcal{H}(\pi(\cdot \mid s_{t})) \right ] 
\end{equation}
where $\alpha$ is the temperature parameter to determine the ratio between the reward and entropy. $\mathcal{H}$ is the entropy that describes the degree of chaos. Then we have new state value function and new loss function of state value. They can be written as follows:
\begin{equation}
V(S_{t})=\mathbb{E}_{a_{t}\sim \pi}\left [ Q(s_{t},a_{t}) \log \pi(a_{t}\mid s_{t})  \right ]
\end{equation}
\begin{equation}
\mathcal{L}_{V}(\psi)=\mathbb{E}_{s_{t}\sim \mu}\left [ \frac{1}{2}(V_{\psi}(s_{t})-\mathbb{E}_{a_{t}\sim \pi_{\phi}}\left [ Q_{\theta}(s_{t},a_{t})-\log \pi_{\phi }(a_{t}\mid s_{t})\right ]) ^{2}  \right ]
\end{equation}
where $\psi$ is the parameter of state value function and $\mu$ is the distribution of states and actions. The new action value function and new loss function of action value can be written as follows:
\begin{equation}
Q(s_{t},a_{t})=r(s_{t},a_{t})+\gamma \mathbb{E}_{s_{t+1}\sim \pi}V(s_{t+1})
\end{equation}
\begin{equation}
\mathcal{L}_{Q}(\theta )=\mathbb{E}_{(s_{t},a_{t})\sim \mu}\left [ \frac{1}{2}(Q_{\theta}(s_{t},a_{t})-r(s_{t},a_{t})-\mathbb{E}_{s_{t+1}\sim \pi_{\phi}}\left [ V_{\bar{\psi} }(s_{t+1})\right ]) ^{2}  \right ]
\end{equation}
where $\theta$ is the parameter of action value function and $\bar{\psi}$ is the weight of the target value network. The loss function of policy can be expressed as follows:
\begin{equation}
\mathcal{L}_{\pi}(\phi)=\mathbb{E}_{s_{t}\sim \mu}\left[ \mathrm {D}_{KL}(\pi_{\phi}(\cdot \mid s_{t})\parallel \frac{\exp(Q_{\theta}(s_{t},\cdot ))}{ {\textstyle \sum_{a_{t}}\exp(Q_{\theta}(s_{t},a_{t}))} } )\right]
\end{equation}
where $\phi$ is the parameter of policy and $\mathrm {D}_{KL}(\cdot \parallel \cdot)$ is the Kullback-Leibler Divergence that measures the distance between the two distributions. And we use reparameterization trick to simplify the calculation process and the loss function of policy changes as follows:
\begin{equation}
\mathcal{L}_{\pi}(\phi)=\mathbb{E}_{s_{t}\sim \mu ,\epsilon_{t}\sim \mathcal{N}}\left [ \log{\pi_{\phi}(f_{\phi}(\epsilon_{t};s_{t})}\mid s_{t}) -Q_{\theta}(s_{t},f_{\phi}(\epsilon_{t};s_{t}))\right ]
\end{equation}
where $\epsilon_{t}$ is a noise parameter sampled from standard Gaussian distribution and then the policy is a function with respect to $\epsilon_{t}$. To improve stability,we calculate the smoothing action value instead of the original method. So the loss function of policy changes as follows:
\begin{equation}
\mathcal{L}_{\pi}(\phi)=\mathbb{E}_{s_{t}\sim \mu ,\epsilon_{t}\sim \mathcal{N}}\left [ \log{\pi_{\phi}(f_{\phi}(\epsilon_{t};s_{t})}\mid s_{t}) -\frac{1}{n}\sum_{i=1}^{n}  Q_{\theta}(s_{t},f_{\phi}(\epsilon_{t_{i}};s_{t}))\right ]
\end{equation}
where parameter $n$ decides the number of action value that is used to calculate smoothing action value. Because $\epsilon_{t}$ is sampled from standard Gaussian distribution, we rewrite it as $\epsilon_{t_{i}}$ to express the change of $\epsilon_{t}$. We called this method smoothing-Q algorithm. And the entire algorithm is summarized as Algorithm 2.

\begin{algorithm}[H]
    \SetAlgoLined
	 \caption{smoothing-Q Algorithm}
    \KwIn{M}
    \KwOut{sq value}
	 Initialize sq $\leftarrow$ 0
    \BlankLine
    \For{$i\leftarrow 1$ \KwTo $M$}{
		$\epsilon_{t_{i}}$ $\leftarrow$ $Normal(0,1).sample$\;
		$a_{t_{i}}$ $\leftarrow$ $f_{\phi}(\epsilon_{t_{i}};s_{t})$\;
		$q$ $\leftarrow$ $\min(Q_{\theta}(s_{t},a_{t_{i}}),Q_{\tilde\theta}(s_{t},a_{t_{i}}))$\;
		sq $\leftarrow$ $($ sq $* i + q$ $)$ / (i+1)\;
	 }
	return sq

\end{algorithm}

\begin{algorithm}[H]
    \SetAlgoLined
	 \caption{SAC-HPO}
    \KwIn{E,T,N}
    \KwOut{The policy network and hyper-parameters}
    Initialize policy $\pi_{\phi}$, action value network $q_{\theta}$, $q_{\tilde{\theta}}$, state value network $v_{\psi}$, $v_{\bar{\psi}}$\;
    Initialize environment, replay buffer $\mathcal{B}$\;
    \BlankLine
    \For{$episode\leftarrow 1$ \KwTo $E$}{
		\For{$t\leftarrow 1$ \KwTo $T$}{
			$s_{t}$ $\leftarrow$ environment.state\;
			$a_{t}$ $\leftarrow$ $f_{\phi}(\epsilon_{t};s_{t})$\;
			$s_{t+1}$, $r_{t+1}$  $\leftarrow$  environment.step($a_{t}$)\;
			$\mathcal{B}$  $\leftarrow$  $\mathcal{B}\bigcup \left \{ s_{t}, a_{t}, s_{t+1},r_{t+1} \right \}$\;
			\For {$n \leftarrow 1$ \KwTo $N$}{
				$\tilde{s}_{t} \leftarrow (1-\alpha)^{n} s_{t}+ {\textstyle \sum_{j=1}^{n}} \alpha(1-\alpha)^{n-j}s_{j}$\;
				$\tilde{s}_{t+1} \leftarrow (1-\alpha)^{n} s_{t+1}+ {\textstyle \sum_{j=1}^{n}} \alpha(1-\alpha)^{n-j}s_{j}$\;
				$\tilde{r}_{t+1} \leftarrow (1-\alpha)^{n} r_{t+1}+ {\textstyle \sum_{j=1}^{n}} \alpha(1-\alpha)^{n-j}r_{j}$\;
				$\mathcal{B}$  $\leftarrow$  $\mathcal{B}\bigcup \left \{ \tilde{s}_{t}, a_{t},\tilde{s}_{t+1},\tilde{r}_{t+1} \right \}$\;
			}
			\If{$\mathcal{B}$.size $>=$ batch size}{
				$\psi \leftarrow \psi-\lambda_{V}\nabla_{\psi}\mathcal{L}_{V}(\psi)$\;
				$\theta \leftarrow \theta-\lambda_{Q}\nabla_{\theta}\mathcal{L}_{Q}(\theta)$\;
				$\tilde{\theta} \leftarrow \tilde{\theta}-\lambda_{Q}\nabla_{\tilde{\theta}}\mathcal{L}_{Q}(\tilde{\theta})$\;
				$\phi \leftarrow \phi-\lambda_{\pi}\nabla_{\phi}\left [ \log{\pi_{\phi}(f_{\phi}(\epsilon_{t};s_{t})}\mid s_{t})-sq\right ]$\;
				$\bar{\psi} \leftarrow \tau \psi+(1-\tau)\bar{\psi}$\;
			}
		}
	}
\end{algorithm}

\section{EXPERIMENTS}
The experiments are performed by using PyTorch on PC with CPU AMD R7-4800u 4.2GHz. In this section, we will demonstrate performance of the proposed algorithm by tuning hyper-parameter of LightGBM and CNN. LightGBM is a well-known algorithm based on decision tree. We pay attention to the performance of hyper-parameter compared to other algorithm. The purpose of our experiment is to prove whether our method is effective.
\subsection{Datasets and Search Space}
To evaluate the universality and generalization of the proposed method, we select three classification datasets from UCI repositories, namely breast cancer dataset, crowd-sourced mapping dataset and HTRU2 dataset respectively. For each dataset, we divide it into two parts for training and validating.
We choose to optimize the hyper-parameters of LightGBM and CNN. LightGBM is developed by Microsoft and it has excellent training efficiency and high accuracy. Five hyper-parameters of LightGBM and CNN are chosen to be optimized respectively. The concrete details are shown in Table 1.

\begin{center}
	\begin{table}[!htbp]\scriptsize
		\caption{Hyper-parameters and losses of our method and Bayesian optimization for LightGBM and CNN} 
		\centering
		\begin{tabular}{lcc|lcc}
		\hline
                 \multicolumn{3}{c|}{LightGBM} & \multicolumn{3}{|c}{CNN} \\
		\hline
	        Hyper-parameter   & Lower             & Upper                    &Hyper-parameter                 & Lower             & Upper \\
		\hline
	        feature fraction & 0.00001           & 1                        & convolution channel           & 1                 & 10 \\
	        learning rate    & 0.00001           & 1                        & convolution kernel            & 1                 & 5 \\
	        bagging fraction & 0.00001           & 1                        & convolution stride            & 1                 & 5 \\
	        reg alpha        & 0                 & 1000                     & fc layer nodes                & 10                & 1000      \\
	        reg lambda       & 0                 & 1000                     & learning rate                 & 0.00001           & 1   \\
        \hline             
       \end{tabular}
	\end{table}
\end{center}
\subsection{Experiment Details}
The critic network, which is the action value network mentioned above, consists of 4 dense layers with 13, 256, 256, 1 nodes respectively. The state value network is the same as critic network basically, except for the input layer that  has 8 nodes because it does not contain actions. The actor network, which is the policy network mentioned above, is the same as state value network except for the output layer. The output layer has 10 nodes that represent mean and variance. The critic network and state value network have the same learning rate 0.0003 but the leaning rate of actor network is 0.003.
In our experiments, we compare our method with Bayesian optimization on the three datasets. Bayesian optimization is a method based on Bayesian method to search for global extreme values of functions, especially high-dimensional nonlinear non-convex functions.

\subsection{Results and Discussion}
The average rewards of our proposed method are presented in Figure 2. We can clearly see the average rewards arise sharply in former 50 episodes. For each dataset, we train it with four methods, which is our proposed method, the method called SQ-HPO removed hierarchical mixture regularization and the method called HMR-HPO removed smoothing-Q. The base method is removed hierarchical mixture regularization and smoothing-Q. We can clearly see our proposed method perform better than the other three. Especially in the early stage, our proposed method ascends faster. For the former dozens of episodes, there is no actual learning because the number of elements in replay buffer is smaller than batch size, which is not enough to train. In this stage, agent can explore randomly so as to cover more hyper-parameters.

\begin{figure}[!htbp]
\centering
\subfigure[\scriptsize LightGBM on Breast Cancer]{
\begin{minipage}[t]{0.5\linewidth}
\centering
\includegraphics[width=2.5in,height=2.1in]{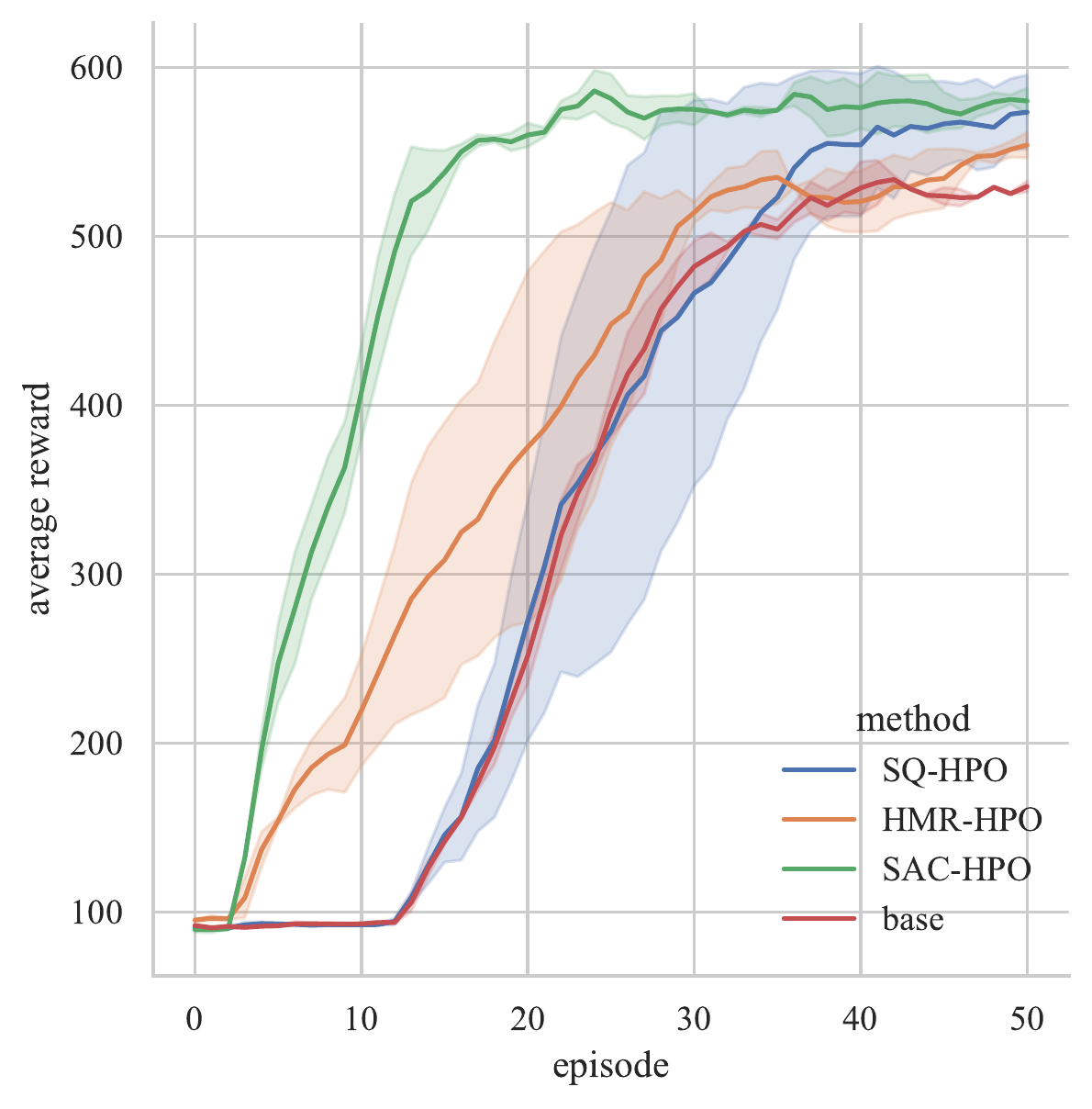}
\end{minipage}
}
\subfigure[\scriptsize LightGBM on Crowd-Sourced Mapping]{
\begin{minipage}[t]{0.45\linewidth}
\centering
\includegraphics[width=2.5in,height=2.1in]{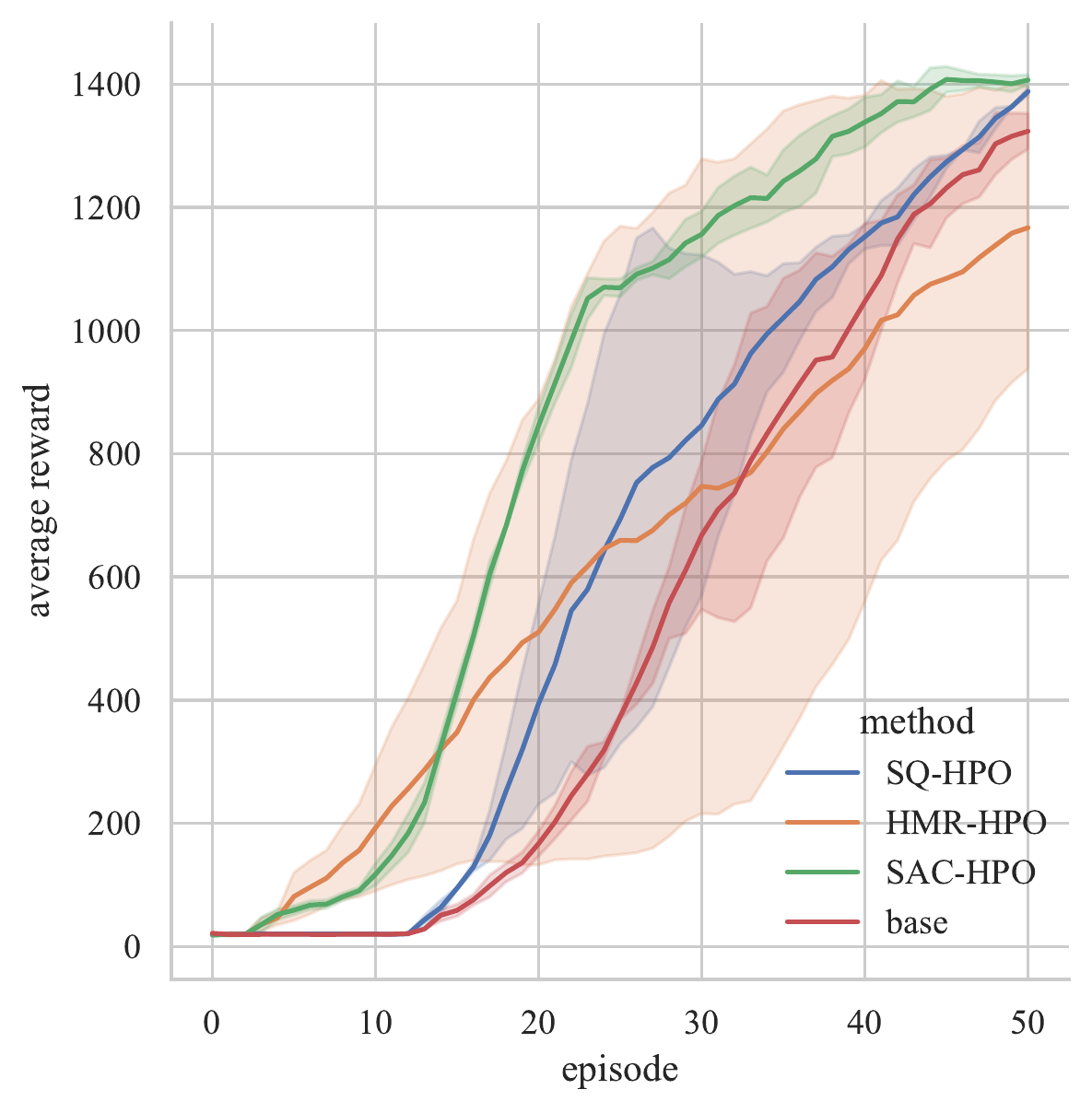}
\end{minipage}
}

\subfigure[\scriptsize LightGBM on HTRU2]{
\begin{minipage}[t]{0,5\linewidth}
\centering
\includegraphics[width=2.5in,height=2.1in]{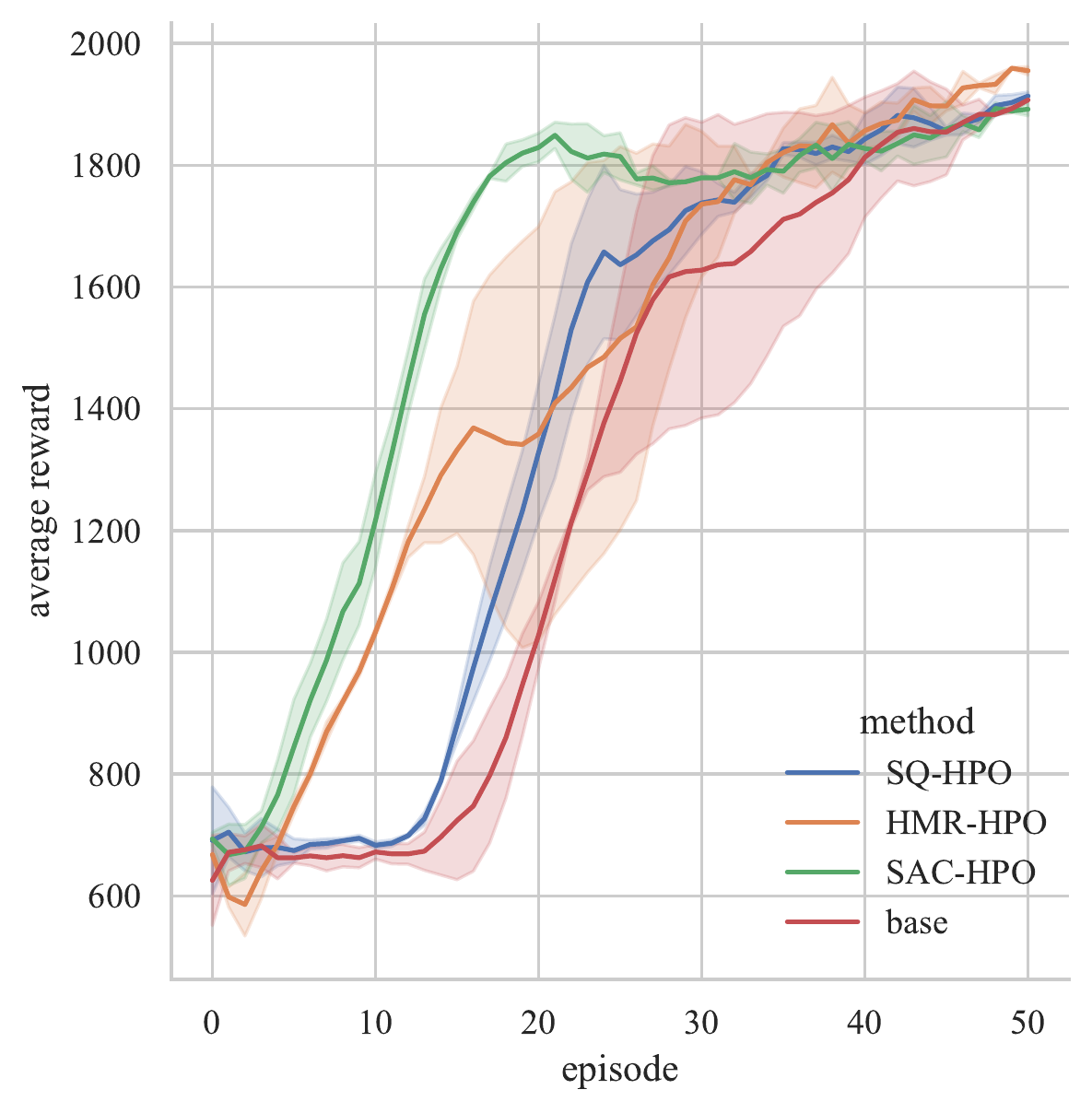}
\end{minipage}
}
\subfigure[\scriptsize CNN on Breast Cancer]{
\begin{minipage}[t]{0.45\linewidth}
\centering
\includegraphics[width=2.5in,height=2.1in]{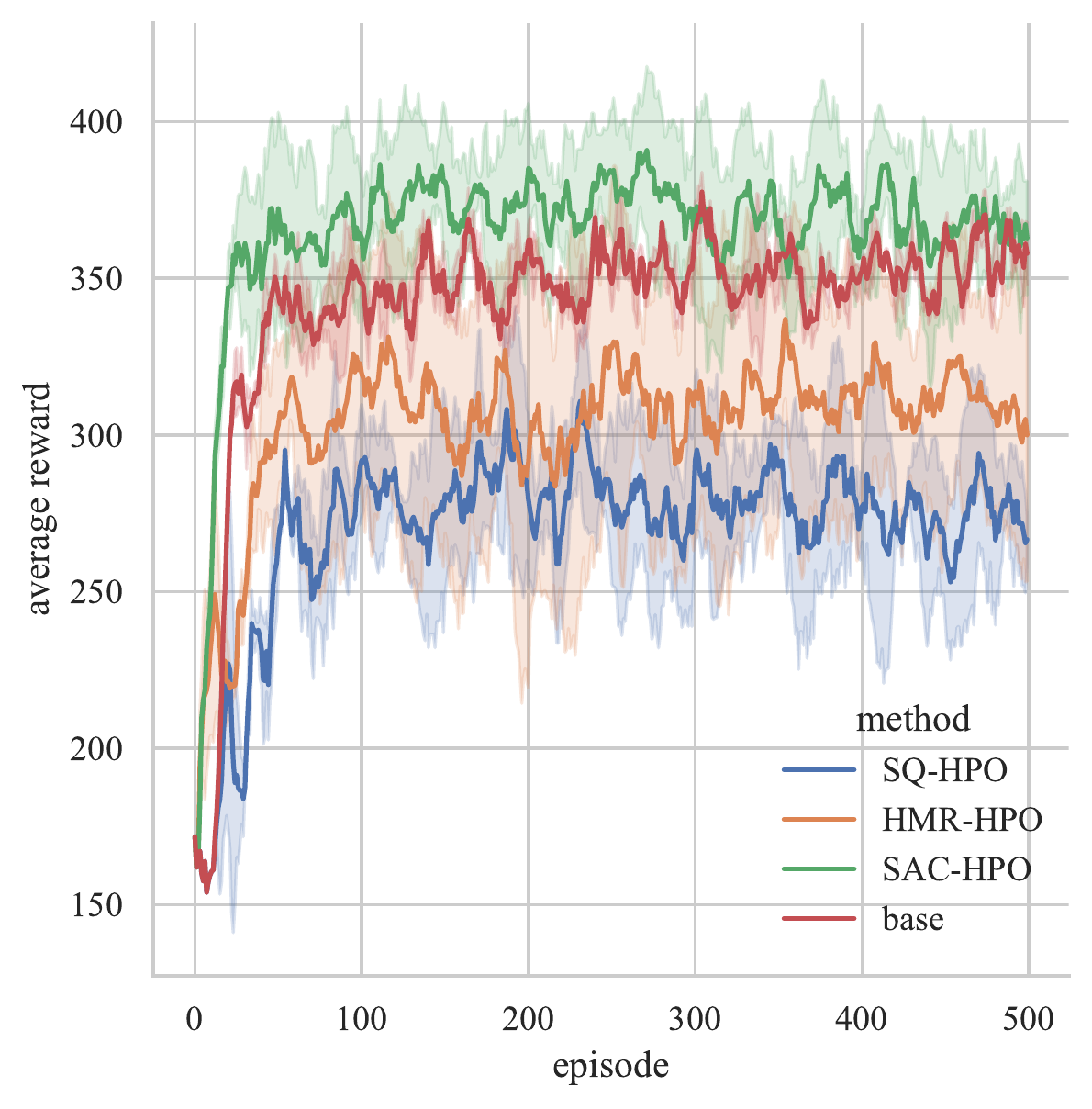}
\end{minipage}
}

\subfigure[\scriptsize CNN on Crowd-Sourced Mapping]{
\begin{minipage}[t]{0,5\linewidth}
\centering
\includegraphics[width=2.5in,height=2.1in]{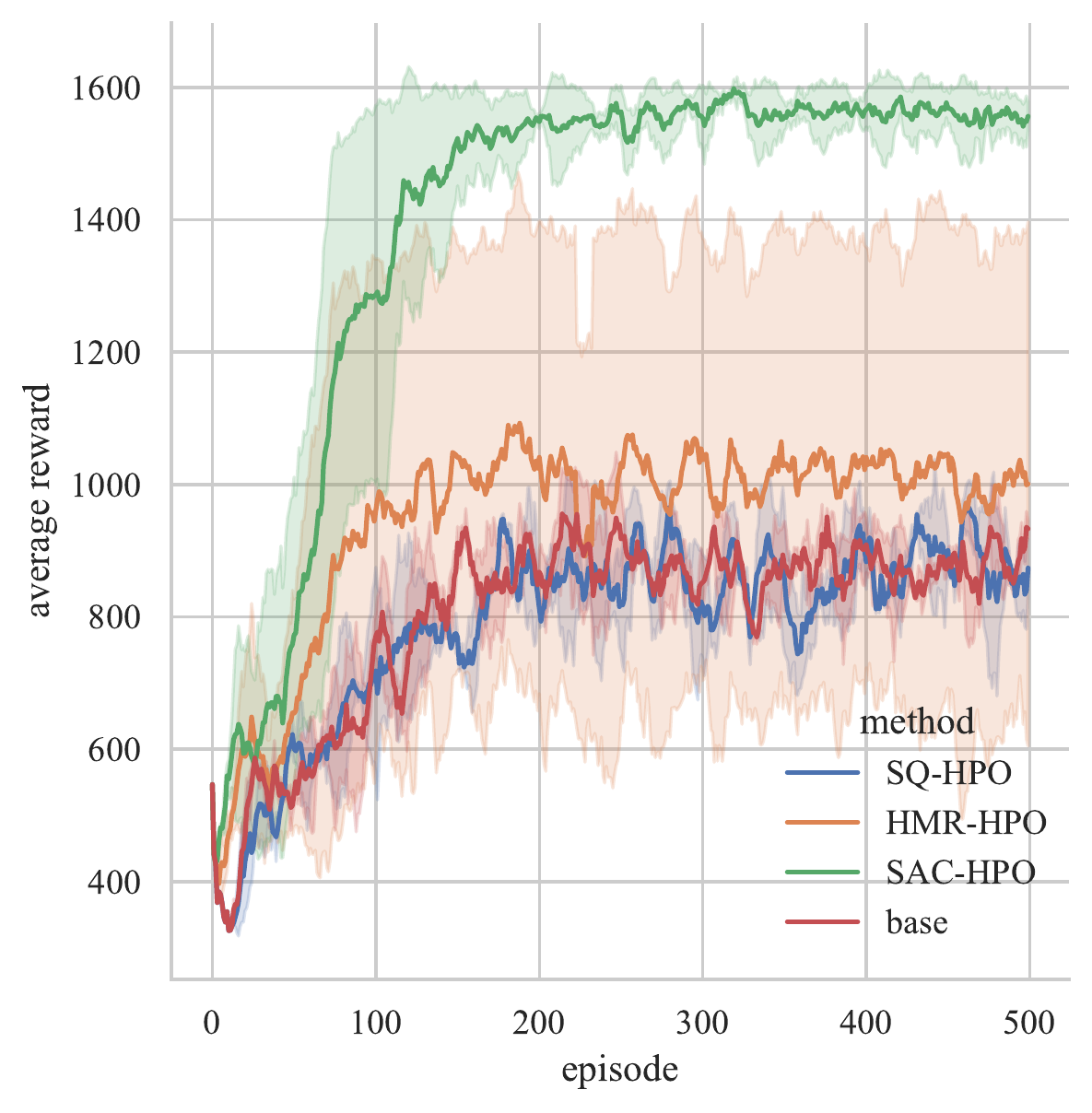}
\end{minipage}
}
\subfigure[\scriptsize CNN on HTRU2]{
\begin{minipage}[t]{0.45\linewidth}
\centering
\includegraphics[width=2.5in,height=2.1in]{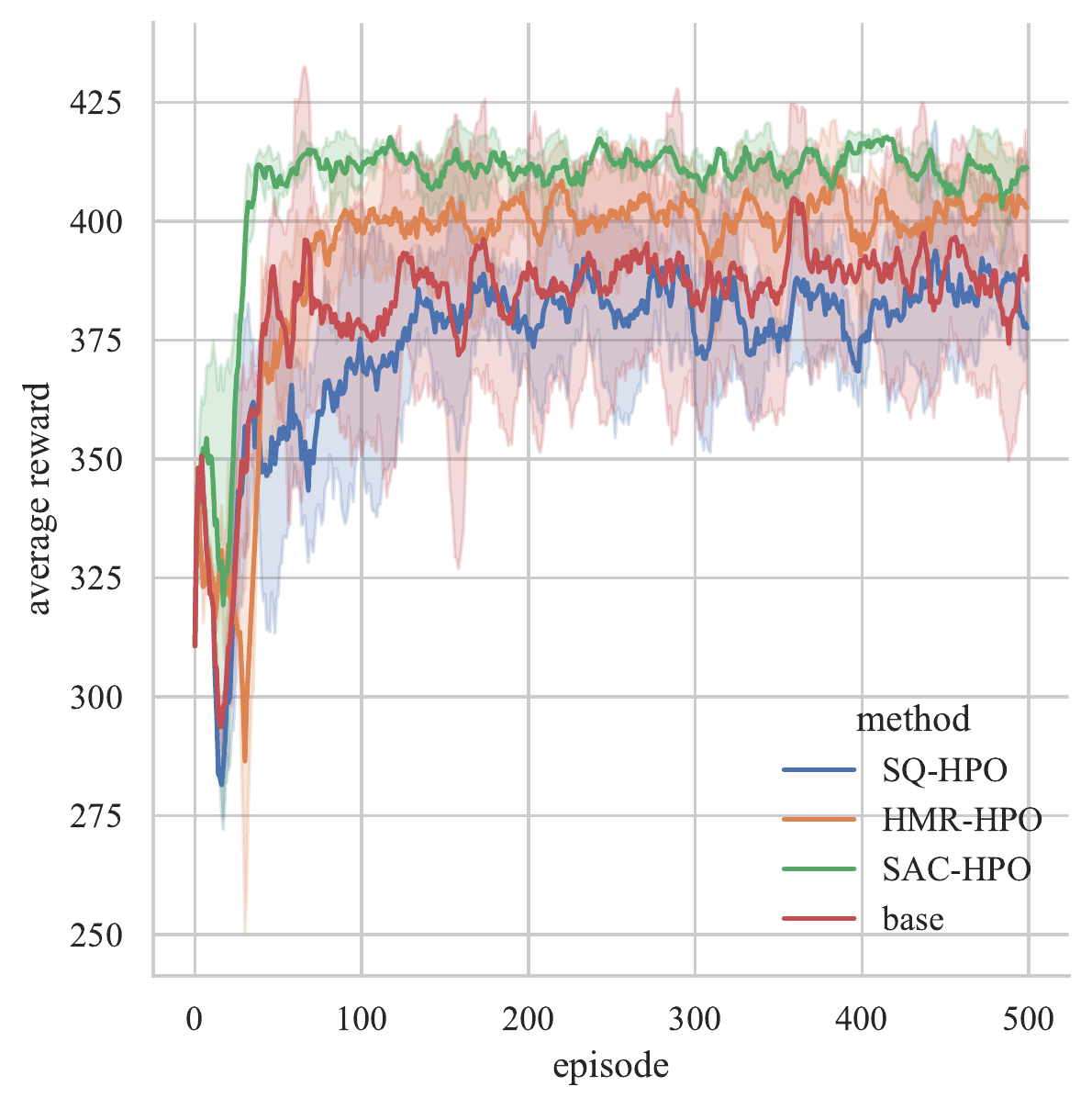}
\end{minipage}
}
\caption{Average reward on three datasets}
\end{figure}

\begin{center}
	\begin{table}[!htbp]\scriptsize
		\caption{Hyper-parameters and losses of our method and Bayesian optimization for LightGBM} \centering
		\begin{tabular}{l|ll|ll|ll}
		\hline
        Dataset          & \multicolumn{2}{|l|}{Breast Cancer} & \multicolumn{2}{|l|}{Crowd-sourced Mapping} & \multicolumn{2}{|l}{HTRU2}  \\
		\hline
        Method           & BO                & SAC-HPO       & BO                & SAC-HPO              & BO                & SAC-HPO \\
        \hline
        feature fraction & 1.0               & 0.1858        & 0.00001           & 0.8485               & 0.6045            & 0.9531 \\
        learning rate    & 1.0               & 0.1476        & 1.0               & 0.8676               & 0.5299            & 0.9542 \\
        bagging fraction & 0.00001           & 0.1757        & 1.0               & 0.7153               & 0.9076            & 0.9530 \\
        reg alpha        & 0                 & 0             & 0                 & 0                    & 3.4748            & 0      \\
        reg lambda       & 881.0848          & 0             & 28.3865           & 0                    & 904.9676          & 1000   \\
        \hline
        mean squared error& 0.0321            &\textbf{0.0242}& 0                 & 0                    & 0.0191            & \textbf{0.0188} \\
        \hline             
       \end{tabular}
	\end{table}
\end{center}
\begin{center}
	\begin{table}[!htbp]\scriptsize
		\caption{Hyper-parameters and losses of our method and Bayesian optimization for CNN} \centering
		\begin{tabular}{l|ll|ll|ll}
		\hline
        Dataset             & \multicolumn{2}{|l|}{Breast Cancer} & \multicolumn{2}{|l|}{Crowd-sourced Mapping} & \multicolumn{2}{|l}{HTRU2}  \\
		\hline
        Method              & BO                & SAC-HPO       & BO                & SAC-HPO              & BO                & SAC-HPO \\
        \hline
        convolution channel & 5                 & 8             & 6                 & 8                    & 2                 & 5 \\ 
        convolution kernel  & 1                 & 1             & 4                 & 5                    & 1                 & 4 \\
        convolution stride  & 4                 & 5             & 2                 & 5                    & 4                 & 4 \\
        fc layer nodes      & 413               & 628           & 583               & 628                  & 72                & 421      \\
        learning rate       & 0.4284            & 0.9638        & 0.3254            & 0.5944               & 0.3607            & 0.7932   \\
        \hline
        cross entropy loss  & 1.7828            & 1.7828        & 2.3005            & \textbf{2.2009}      & 1.5607            & \textbf{1.5549} \\
        \hline             
       \end{tabular}
	\end{table}
\end{center}

We compare our method with Bayesian optimization in three datasets and select best hyper-parameters of each dataset. The concrete details are shown in Table 2 and Table 3. Our method achieves the same performance as Bayesian optimization or better. We can explicitly see that our method and Bayesian optimization achieve the same loss on crowd-sourced mapping dataset for LightGBM and breast cancer dataset for CNN, but the hyper-parameters are different. In other datasets, our method performs better than Bayesian optimization.
\section{CONCLUSIONS}
In this paper, a new hyper-parameter optimization method based on reinforcement learning with hierarchical mixture regularization and smoothing-Q is proposed. %This paper lays the foundation for using soft actor critic method to control hyper-parameter optimization. 
This method utilizes LSTM as the environment interacted with the agent which generates the hyper-parameters automatically for machine learning models on a given dataset. Reward function based on the loss of validation set is also proposed. The hierarchical mixture regularization is used to deal with the problem of low sample efficiency, which is caused by the high computational cost of the loss. In order to verify the feasibility of our method, we conducted a series of experiments of tuning hyper-parameters for LightGBM and CNN. The results of the experiments prove that soft actor critic with hierarchical mixture regularization and smoothing-Q outperforms original soft actor critic method.

\section*{Acknowledgement}
This work is supported by NSFC-61803337, NSFC-61803338.

\bibliographystyle{elsarticle-num}
\bibliography{reference}

\end{document}